# LLM-IE: A Python Package for Generative Information Extraction with Large Language Models

Enshuo Hsu[1,2], MS; Kirk Roberts[1], PhD

[1]McWilliams School of Biomedical Informatics, University of Texas Health Science Center at Houston, Houston, Texas, USA.

[2]Enteprise Development and Integration, University of Texas MD Anderson Cancer Center, Houston, TX, USA.

**Corresponding author: Kirk Roberts, kirk.roberts@uth.tmc.edu**

## ABSTRACT

**Objectives**

Despite the recent adoption of large language models (LLMs) for biomedical information extraction, challenges in prompt engineering and algorithms persist, with no dedicated software available. To address this, we developed *LLM-IE*: a Python package for building complete information extraction pipelines. Our key innovation is an interactive LLM agent to support schema definition and prompt design.

**Materials and Methods**

The *LLM-IE* supports named entity recognition, entity attribute extraction, and relation extraction tasks. We benchmarked on the i2b2 datasets and conducted a system evaluation.

**Results**

The sentence-based prompting algorithm resulted in the best performance while requiring a longer inference time. System evaluation provided intuitive visualization.

**Discussion**

*LLM-IE* was designed from practical NLP experience in healthcare and has been adopted in internal projects. It should hold great value to the biomedical NLP community.

**Conclusion**

We developed a Python package, *LLM-IE*, that provides building blocks for robust information extraction pipeline construction.

## BACKGROUND AND SIGNIFICANCE

The use of large language models (LLMs) for information extraction in natural language processing (NLP) has gained increasing popularity [1]. There are several benefits including 1) low annotation requirement through zero-shot and few-shot learning [2,3], 2) comparable performance to fully fine-tuned models [3], and 3) end-to-end entity span and relation extraction [4]. In the biomedical field where manually labeled gold standards are expensive and information extraction schemas are often complex, LLM-based information extraction methods show great promise. Recent works have been focusing on 1) LLM inferencing infrastructures [5–7], 2) LLM prompting algorithms [3,4,8–14], and 3) prompt engineering [15]. However, for NLP practitioners, challenges persist as the inference engines are difficult to configure and depend heavily on computing environment. Further, prompt engineering requires experience, domain knowledge, and effort in iterative development. Finally, despite some studies releasing source code, to our knowledge no software integrates multiple systems and methods and provides a comprehensive toolkit for the LLM-based information extraction pipeline building. Therefore, we developed a Python package, *LLM-IE*, for the clinical NLP community.

Our work has the following significance:

1. We provide a uniform interface for different LLM inference engines which avoids the complexity of configuration.
2. We implement popular prompting algorithms published in the biomedical domain and the open domain and provide simple APIs.
3. We build an LLM agent ("Prompt Editor") to help users write and polish prompt templates.

## OBJECTIVE

We published a Python package on the Python Package Index (PyPi) repository and the GitHub repository.

## METHODS

*LLM-IE* is a comprehensive toolkit that provides building blocks for the construction of LLM-based information extraction pipelines. The package and documentation are available on PyPi (https://pypi.org/project/llm-ie/ ) and GitHub (https://github.com/daviden1013/llm-ie )

### Usage

*LLM-IE* covers the life cycle of an NLP information extraction pipeline: 1) task definition, 2) prompt design, 3) named entity extraction, 4) entity attributes extraction, 5) relation extraction, 6) data management, and 7) visualization.

In the **task definition** and **prompt design** phases, users work closely with the Prompt Editor, is an LLM agent with access to many pre-stored prompt templates and guidelines. Users choose an information extraction algorithm ("extractor") and start chatting with the Prompt Editor via terminal or IPython (e.g., Jupyter Notebooks). On the backend, the Prompt Editor analyzes the users' requests using the relevant templates and prompt-writing guidelines and generates a prompt template with specific task descriptions, schema definition, output format definition, and input placeholders.

The system prompt for the Prompt Editor:

> You are an AI assistant specializing in prompt writing and improvement. Your role is to help users refine, rewrite, and generate effective prompts based on guidelines provided…

The chat prompt template includes a placeholder for prompt guidelines and examples:

> # Task description
>
> Chat with the user following the prompt guideline below.
>
> # Prompt guideline
>
> {{prompt_guideline}}

Users are encouraged to iteratively develop with the Prompt Editor until a final prompt template is prepared. In the **named entity extraction** and **entity attributes extraction** phases, the frame extractor applies the prompt template for end-to-end entity spans and attribute extraction on the target documents. The LLM outputs strings following the JSON schema specified in the prompt template. A post-processing method then converts them into structured frames with frame ID, entity text, entity spans, and a set of attributes. The **relation extraction** phase involves the extracted frames from the previous step and a relation extraction prompt template which can be constructed by working with the Prompt Editor. The relation extractors apply the prompt template on pairs of frames to detect relation existence (i.e., binary relations) and relation types (i.e., multi-class relations). To reduce computation for LLM inferencing, users are encouraged to provide a pre-processing function (i.e., possible_relation_types_func) that applies decision rules. For example, if the two frames in the pair are "drug" and "dosage", the possible relation types are "Dosage-Drug" and "No-relation", while "dosage" and "dosage" frames must be "No-relation" and thus do not require LLM inferencing. After extraction, the built-in data types (e.g., LLMInformationExtractionDocument) process, store, and visualize the frames and relations via a Flask App or HTML rendering (Figure 1).

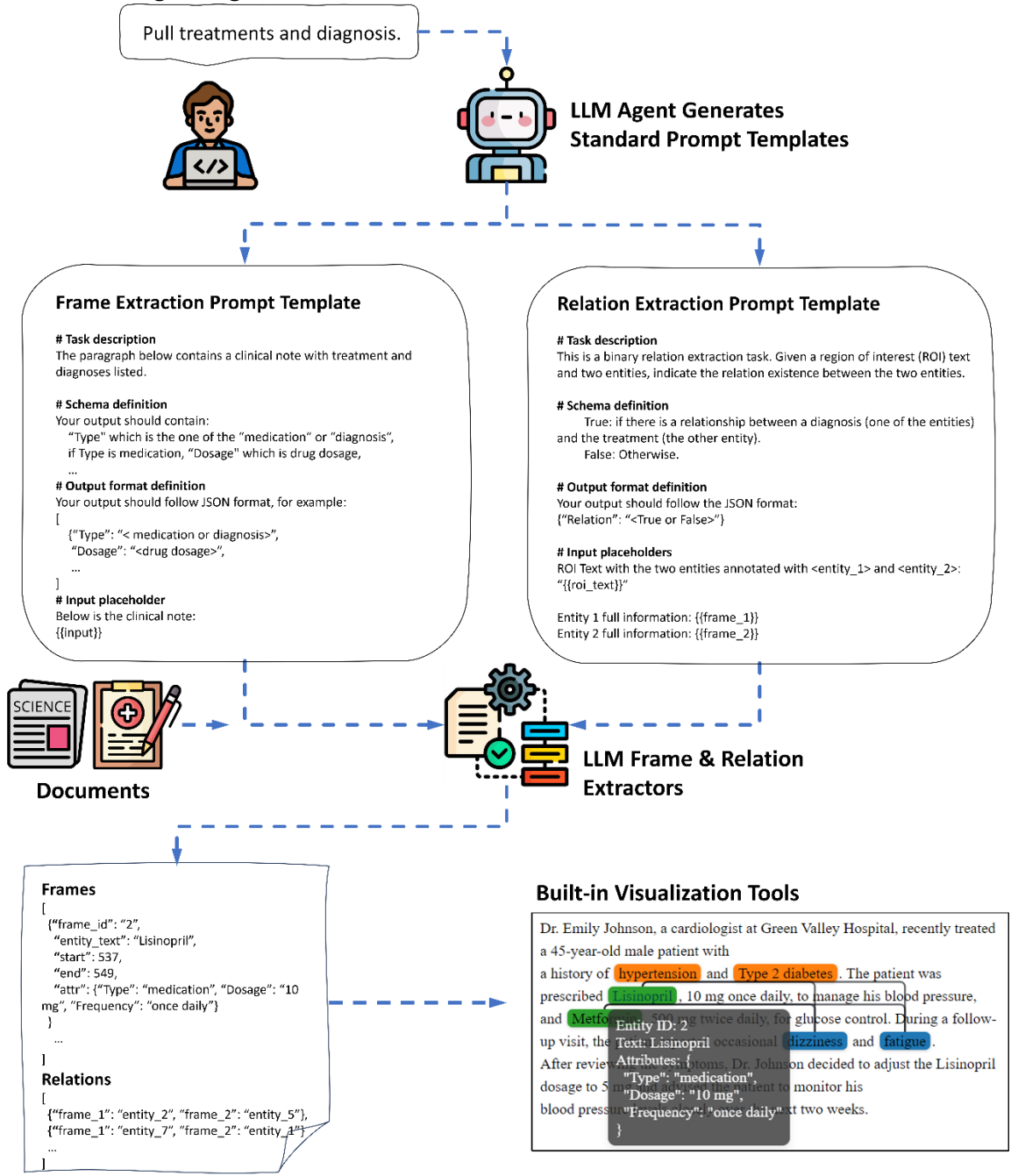

Figure 1: Usage flowchart. Users start by providing a simple description of the task to the LLM agent. The LLM agent generates standard prompt templates with Task description, schema definition, output format definition, and input placeholders. Users iteratively develop prompt templates with the LLM agent until a high-quality prompt template is prepared. The FrameExtractors use the prompt template to extract entities and attributes ("frames"). The RelationExtractors extract the relation and relation types between frames. The built-in visualization tools render the frames and relations on a web App.

## System Design

Our system design follows four principles: 1) Efficiency, in which recent and successful inference engines and prompting algorithms are supported (e.g., Ollama [5], HuggingFace-hub [16], Llama.cpp [6], vLLM [7], OpenAI API). 2) Flexibility, in which fundamental functions are implemented as modules and classes (e.g., Inference Engines, Frame Extractors, Relation Extractors) for easy customization. 3) Transparency, in which all the prompt templates, LLM inputs, and outputs are accessible to users. 4) Minimalism, in which the package has few dependencies. Users only install dependencies for functions they use.

The LLM-IE package is composed of four Python modules: 1) Engines, 2) Extractors, 3) Data types, and 4) Prompt Editor. The **Engines module** defines interface classes that support popular open-source (e.g., Ollama, HuggingFace-hub) and closed-source (e.g., OpenAI API) LLM inference engines. They work for the Prompt Editor and extractors. The **Extractors module** defines prompting algorithms ("extractors") for frame and relation extraction. The Basic frame extractor prompts LLM directly and outputs a list of frames. The Review frame extractor prompts LLM to generate initial outputs and prompt again for amendment and correction. The Sentence frame extractor splits the target document into sentences and prompts sentence by sentence to improve recall and entity span detection accuracy. The binary relation extractor prompts LLM to review and detect relations between a pair of frames. The multi-class relation extractor prompts LLM to classify relation types between a pair of frames. The algorithm sources are summarized in Table 1. More implementation details are shown in Table SX. The **Data types module** defines data management classes for frames and relations storage, validation, and visualization. A document is packaged into a self-contained object. The validation checks for overlaps and redundancy and ensures that relations are linking two existing frames. For minimalism, we implemented the visualization methods (i.e., viz_serve, viz_render) by internally calling our plug-in Python package, "ie-viz". The **Prompt editor module** defines a Prompt Editor class that serves as an LLM agent for prompt development. It has access to pre-stored prompt-writing guidelines and examples for each extractor (Figure 2).

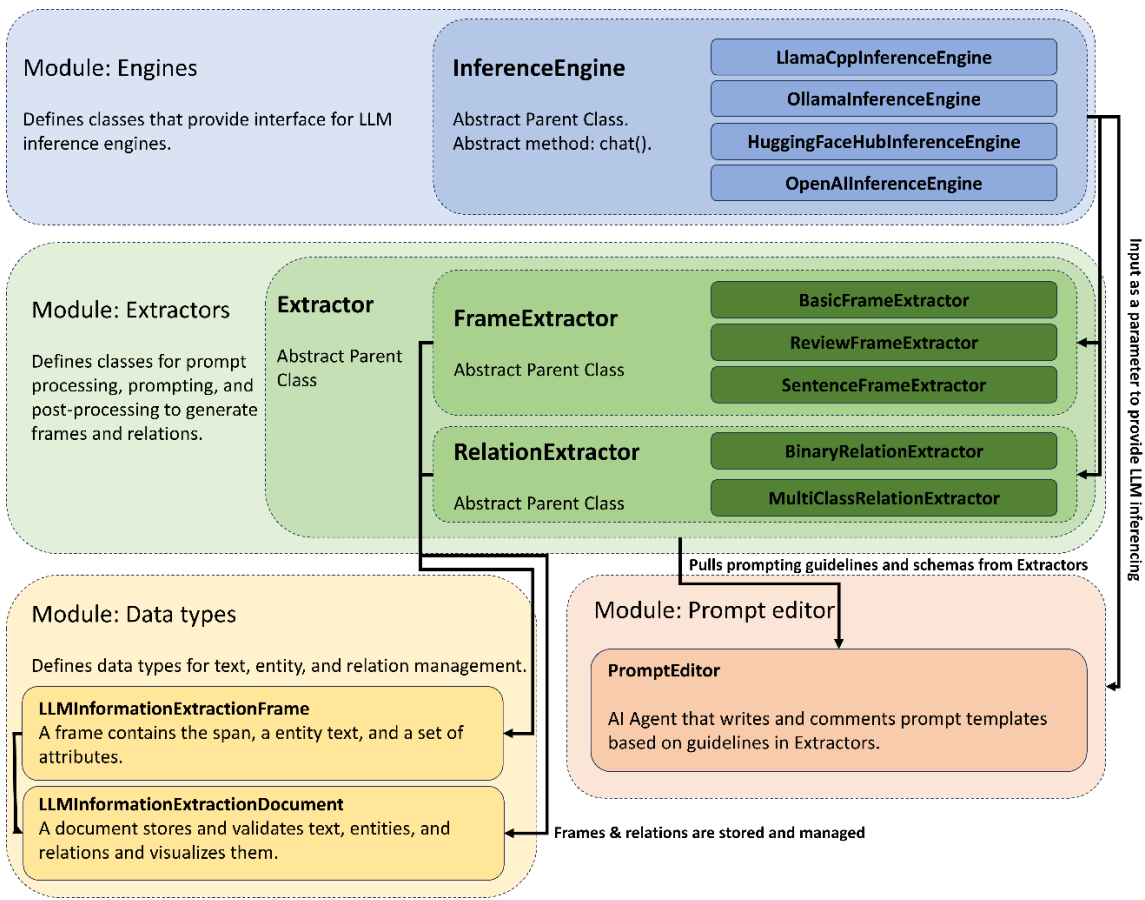

*Figure 2: Conceptual class diagram. The Engines module defines InferenceEngine classes that host LLM and provides an interface for inference. The Extractors module defines FrameExtractors and RelationExtractors that process apply prompt templates, prompt LLM for information extraction, and post-process outputs. The Data types module defines containers for text, entities, and relations management and visualization. The PromptEditor module defines a PromptEditor AI Agent to write and comment on prompt templates.*

*Table 1: Prompting algorithm sources*

| Task | Algorithms (implemented in extractors) | Algorithm references |
|---|---|---|
| Named entity recognition | BasicFrameExtractor<br>ReviewFrameExtractor<br>SentenceFrameExtractor | [3]<br>[8,9]<br>[10,11,17] |
| Entity attribute extraction | All above FrameExtractors | [4] |
| Relation extraction | BinaryRelationExtractor<br>MultiClassRelationExtractor | [12,13]<br>[12–14] |

**Benchmarking and System Evaluation**

We benchmarked our package on three clinical NLP datasets for named entity recognition (NER), entity attribute extraction (EA), and relation extraction (RE). We adopted the 2012 [18] and 2014 [19] Integrating Biology and the Bedside (i2b2), and 2018 National NLP Clinical Challenges (n2c2) [20] Natural Language Processing Challenge. All experiments were evaluated with the Llama-3.1-70B [21] in an 8-shot prompting setting and conducted with the vLLM [7] inference engine on a GPU server with 8 NVIDIA A100 GPUs. Details and source code are discussed on our GitHub page (https://github.com/daviden1013/LLM-IE_Benchmark ).

The i2b2/ n2c2 data user agreement prohibits public sharing of the text content. Therefore, we performed a system evaluation and visualized the extraction on a synthesized clinical note. The task is to extract drugs, conditions, and adverse drug events (ADEs) with corresponding attributes and relations. Implementation details are available on our GitHub page (LLM-IE_Benchmark).

## RESULTS

### Benchmarking

For the NER and EA tasks, the Sentence Frame Extractor achieved the best F1 scores, while consuming more GPU time. The Review Frame Extractor had higher recall than the Basic Frame Extractor on all NER tasks.

*Table 2: Benchmark on the i2b2/ n2c2 datasets for NER, EA, and RE tasks*

| Tasks | Algorithm | GPU time (s)/ Note | Benchmarks | | | | | |
|---|---|---|---|---|---|---|---|---|
| | | | **2012 Temporal Relations Challenge** | | | | | |
| | | | EVENT | | | TIMEX | | |
| | | | Precision | Recall | F1 | Precision | Recall | F1 |
| | Basic | 67.5 | 0.9406 | 0.2841 | 0.4364 | 0.9595 | 0.3516 | 0.5147 |
| | Review | 84.0 | 0.8965 | 0.3995 | 0.5527 | 0.9352 | 0.5473 | 0.6905 |
| | Sentence | 132.9 | 0.9101 | 0.6824 | 0.7799 | 0.8891 | 0.739 | 0.8071 |
| | | | **2014 De-identification Challenge** | | | | | |
| | | | Strict | | | Relaxed | | |
| **Named Entity Recognition** | | | Precision | Recall | F1 | Recall | Precision | F1 |
| | Basic | 9.4 | 0.7154 | 0.4813 | 0.5755 | 0.7172 | 0.4826 | 0.5769 |
| | Review | 15.7 | 0.5649 | 0.5454 | 0.555 | 0.5667 | 0.5471 | 0.5567 |
| | Sentence | 20.7 | 0.6683 | 0.7379 | 0.7014 | 0.6703 | 0.7401 | 0.7035 |
| | | | **2018 (Track 2) ADE and Medication Extraction Challenge** | | | | | |
| | | | Strict | | | Lenient | | |
| | | | Precision | Recall | F1 | Recall | Precision | F1 |
| | Basic | 44.3 | 0.7384 | 0.3534 | 0.478 | 0.8537 | 0.4034 | 0.5479 |
| | Review | 63.2 | 0.7209 | 0.427 | 0.5363 | 0.8416 | 0.4918 | 0.6208 |
| | Sentence | 114.1 | 0.852 | 0.6166 | 0.7154 | 0.963 | 0.692 | 0.8053 |
| | | | **2012 Temporal Relations Challenge** | | | | | |
| **Entity Attribute Extraction** | | | EVENT | | | TIMEX | | |
| | | | Type | Polarity | Modality | Type | Value | Modifier |
| | Basic | 67.5 | 0.2589 | 0.2707 | 0.2737 | 0.3236 | 0.2835 | 0.3198 |
| | Review | 84.0 | 0.358 | 0.3799 | 0.3828 | 0.4934 | 0.4209 | 0.4857 |
| | Sentence | 132.9 | 0.6056 | 0.642 | 0.6432 | 0.678 | 0.5505 | 0.667 |
| **Relation Extraction** | | | **2018 (Track 2) ADE and Medication Extraction Challenge** | | | | | |
| | | | Precision | | Recall | | F1 | |
| | Multi-class | 213.9 | 0.3831 | | 0.978 | | 0.5505 | |

### System Evaluation

We utilized the LLE-IE package to build an information extraction pipeline for the drug, condition, and ADE entities, attributes, and relations. For all the frames extracted by the Frame extractor, the attribute "Type" represents the frame type as one of the "Drug", "Condition", or "ADE". If the Type is "Drug", "Dosage" and "Frequency" are extracted as additional attributes. If the Type is "Condition", an "Assertion" attribute is assigned. The relations between a "Condition" frame and a "Drug" frame and between an "ADE" frame and a "Drug" frame are extracted by the Relation extractor. We visualized the results with the viz_render method and displayed them on a browser (Figure 3).

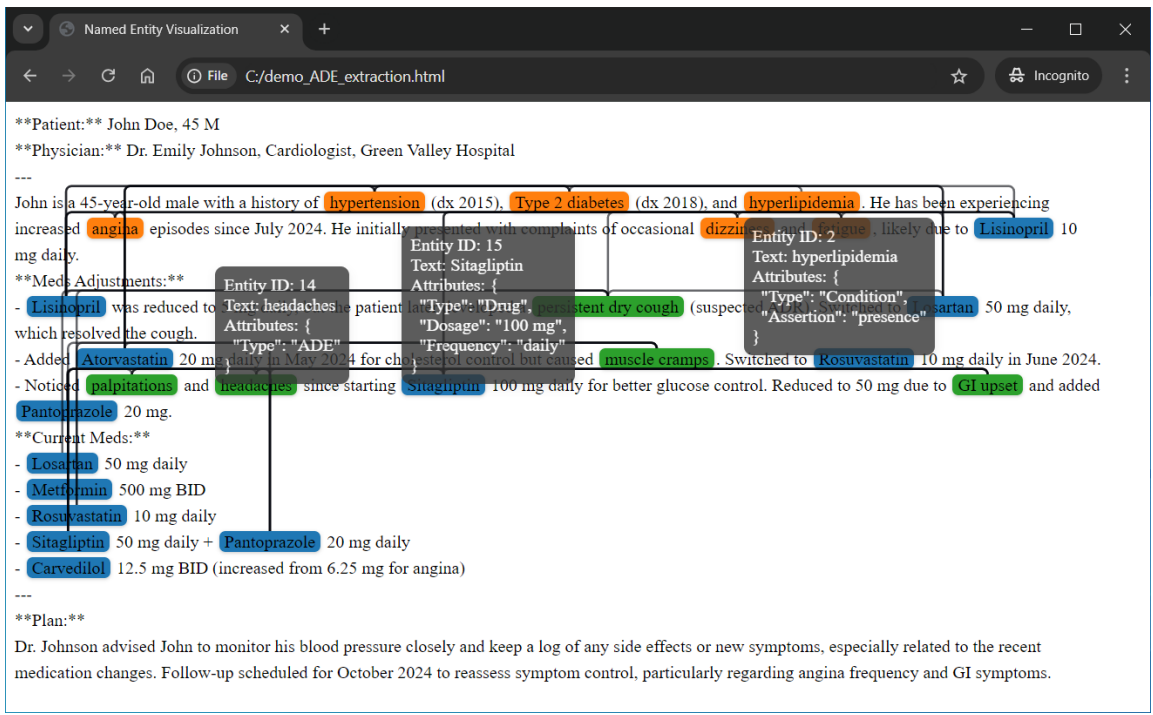

*Figure 3: System performance and visualization. The frames are highlighted based on the attribute "Type" as Drug, Condition, or ADE. For the Drug frames, attributes "Dosage" and "Frequency" are extracted. For the Condition frames, the attribute "Assertion" is extracted. The relations Condition-Drug and ADE-Drug are visualized as paths. Note that for publication purposes, only a few entity attributes are displayed in this figure.*

## DISCUSSION

We developed the *LLM-IE* Python package for LLM-based information extraction. The usage (i.e., building block classes and pipelines) is designed based on our practical NLP experience in the healthcare industry. We have been adopting it internally for NLP projects. Therefore, we believe it is relevant to other NLP practitioners in the biomedical field. The system design in which inference engines and extractors are placed in modules with well-organized inherent relationships enables continuous development as new infrastructures and prompting algorithms are released in the future. Our visualization features provide an intuitive way to validate (e.g., error analysis, performance evaluation) outputs with a complex schema which would be cumbersome otherwise.

## CONCLUSIONS

To fill in the gaps between the latest LLM technology and biomedical NLP practices, we developed a

The benchmark results are reasonable compared to our recent publication [22]. In some cases, the few-shot LLM performance was below fully supervised models, as previously reported [15].

Despite the great features, our *LLM-IE* package has a few limitations: 1) it is in an active development phase. More practical adoption and evaluation are needed. 2) Like all LLM-based systems, prompt engineering plays an important role in providing domain knowledge and task-specific definitions. Despite our Prompt Editor LLM agent, it is up to the users to finalize the prompt templates. Some familiarity with prompt writing is still necessary. 3) The post-processing relies on the LLM to output in the correct format. Inconsistent elements in the JSON list are discarded. Thus, it is important to choose instructed LLMs with good instruction-following performance. 4) Our benchmarking and system evaluation used Llama 3.1 to represent the state-of-the-art open-source LLM at this point. Further evaluation is needed for other LLMs.

Python package, *LLM-IE*, that provides building blocks for robust information extraction pipeline construction.